\documentclass[11pt]{article}

\usepackage[a4paper,margin=2.4cm]{geometry}
\usepackage{fontspec}
\newfontfamily\arabicfont{Amiri-Regular.ttf}[Path=fonts/,BoldFont=Amiri-Bold.ttf,Script=Arabic,Scale=1.08]
\newcommand{\ar}[1]{{\arabicfont\beginR #1\endR}}
\newcommand{\num}[1]{{\beginL #1\endL}}

\usepackage[numbers,sort&compress]{natbib}
\usepackage{booktabs}
\usepackage{array}
\usepackage{tabularx}
\usepackage{xcolor}
\usepackage{tikz}
\usetikzlibrary{arrows.meta,positioning,shapes.misc,fit}
\usepackage{amsmath}
\usepackage{enumitem}
\usepackage{caption}
\usepackage{microtype}
\usepackage[hidelinks]{hyperref}
\usepackage{url}

\definecolor{navy}{HTML}{1F3864}
\definecolor{sand}{HTML}{F3EEE4}
\setlist{nosep,leftmargin=1.4em}
\newcommand{\qistas}{Qistas}
\newcommand{\mizan}{Mizan}

\title{\textbf{\qistas{}: A Pre-Registered, Portable Protocol for Fairness Evaluation of Large Language Models}\\[4pt]
\large Design, Tooling, and a Reviewed Seed for the First Iraqi Instantiation}

\author{
Nawar S. Alseelawi\textsuperscript{1} \and Mustafa S. Aljumaily\textsuperscript{2}\\[6pt]
\normalsize \textsuperscript{1}University of Misan, Amarah, Iraq \quad \textsuperscript{2}Missan Oil Company, Amarah, Iraq\\
\normalsize \texttt{nawar.alseelawi@uomisan.edu.iq} \quad \texttt{mustafasadiq0@gmail.com}
}
\date{October 2026}

\begin{document}
\maketitle

\begin{abstract}
Fairness benchmarks for large language models (LLMs) are overwhelmingly English-born and American in their social categories; translated versions carry those categories into societies where they do not apply, and almost none fix their hypotheses before the models are run. We present \qistas{} (Arabic \ar{قِسطاس}, \emph{qis\d{t}\=as}, ``the just balance''), a protocol for evaluating how LLMs treat the social, confessional, and regional constituencies of a society, designed to be portable across countries and instantiated first for Iraq. The protocol rests on a deliberate restriction: in its pilot form it measures only what does not require a verdict on contested truth - procedural symmetry of treatment across matched counterfactual pairs, differential refusal, stereotype reliance under ambiguity, and terminological framing of documented events. Every measurement is paired by construction and analysed with paired machinery (sign-flip permutation tests, exact McNemar tests, template-clustered bootstrap intervals, Holm-corrected families), with a power analysis that states, before any data exist, the smallest effect each axis can detect once the design effect of template clustering is applied. We release the authoring apparatus - a pruned $3\times5$ taxonomy with four cells closed by stated design, per-type authoring checklists, a cumulative rulings log, an automated structural audit, and a human review notebook - and report its first field test: a semantic review of a 58-item seed bank in which 24 of 25 template groups failed at least one checklist item (36 failed checks in all). The defect patterns that review exposed (evaluation criteria leaking into prompts, token-level swaps that break Arabic agreement, epistemically unbalanced pairs, single-theme refusal probes, unwritten dialect twins counted as coverage) are, we argue, general hazards of counterfactual fairness benchmarks rather than local accidents. This paper reports a protocol and the validation of its apparatus, not model results: no model has been evaluated, and the pre-registration published alongside it commits the hypotheses, metrics, exclusion rules, and analysis plan that the forthcoming pilot will follow.
\end{abstract}

\section{Introduction}

The question ``does this model treat people fairly?'' has, in practice, meant ``does this model reproduce the stereotypes catalogued in a handful of English datasets?'' BBQ \citep{parrish2022bbq}, CrowS-Pairs \citep{nangia2020crows}, StereoSet \citep{nadeem2021stereoset}, and HolisticBias \citep{smith2022holisticbias} shaped how the field measures social bias, and each encodes a particular society's categories - race as the United States counts it, religion as a Western census lists it, region as American geography draws it. When such datasets travel by translation, the categories travel with them, and the result measures how a model handles a foreign taxonomy rather than how it treats the people who will actually use it. A second generation of country-specific benchmarks - French CrowS-Pairs \citep{neveol2022frenchcrows}, KoBBQ for Korea \citep{jin2024kobbq}, CBBQ for China \citep{huang2024cbbq}, BharatBBQ for India \citep{tomar2025bharatbbq}, and MBBQ across Dutch, Spanish, and Turkish \citep{neplenbroek2024mbbq} - shows that the field has recognised the problem; each rebuilds the categories for its society, and KoBBQ in particular documents how much of the original BBQ did not survive the crossing. The pitfalls that persist are documented in detail \citep{blodgett2021salmon,blodgett2020power}: unclear constructs, incoherent pairs, stereotypes that are not stereotypes anywhere in particular, and an absence of any statement of what a difference between two sentences is supposed to show.

Arabic LLM evaluation has begun to take culture seriously \citep{naous2024beer,myung2024blend,chiu2025culturalbench}, but these efforts measure cultural \emph{knowledge} and \emph{adaptation} - whether a model knows that Iraqis drink tea in \ar{استكان} (\emph{istik\=an}) glasses, not whether it describes a job applicant from Nineveh with the same depth and tone it affords one from Basra. Our earlier work on \mizan{} \citep{alseelawi2026mizan}, a national benchmark for Iraqi Arabic and the Iraqi civic context, exposed the gap from the other side: the benchmark was built to measure competence, yet its most consequential finding concerned conduct - a safety-hardened model family deterministically refused innocuous dialect-comprehension items as policy violations. Differential refusal is a fairness phenomenon, and a competence benchmark found it only by accident. \qistas{} is the deliberate instrument.

Three commitments distinguish the protocol. First, \textbf{procedural neutrality}: \qistas{} never asks whether a model's answer is \emph{true} about a contested matter and never ranks the constituencies it covers; it asks whether a model's treatment of two matched prompts is \emph{symmetric}. The pilot scope is cut to what this commitment permits - the social-regional, confessional-communal, and historical-political domains of Iraq, with the last restricted to verifiable facts and to the vocabulary models choose when describing documented events. Second, \textbf{paired design and paired statistics}: the unit of measurement is the counterfactual pair or template group, never the single item, and every reported number comes with an interval computed under the dependence that templates induce. Third, \textbf{pre-registration}: the hypotheses, metrics, exclusion rules, and correction families are fixed and published before any model is run, following the practice argued for in NLP by \citet{vanmiltenburg2021prereg} and in the sciences generally by \citet{nosek2018prereg}.

The paper is a protocol paper in the sense of a registered design: it reports what will be measured and how, and it reports the validation of the apparatus, not the measurements. We make this explicit because the field rarely does. Its contributions are:

\begin{enumerate}
\item A portable protocol built on a pruned taxonomy (three domains $\times$ five manifestation forms, eleven active cells and four closed by design with stated reasons) and four paired test types, each with an explicit authoring guide and reviewer checklist, together with an instantiation checklist that says what a second country must redo (\S\ref{sec:taxonomy}, \S\ref{sec:authoring}, Appendix~\ref{app:instantiation}).
\item A binding statistical protocol with a computed power analysis that states, per axis and after the design effect of clustering, the minimum detectable effect at the planned bank size, and a declared rule that confirmatory analysis is axis-level while single-domain slices are descriptive (\S\ref{sec:stats}).
\item An authoring and calibration apparatus - automated structural audit, cumulative rulings log, and a human review notebook - together with the results of its first field test on a 58-item seed bank, which surfaced defect patterns we believe generalise beyond Iraq (\S\ref{sec:calibration}).
\item A public pre-registration of four confirmatory hypotheses with their test statistics, families, and replication rule, and a list of what will \emph{not} be tested confirmatorily, plus an open reference implementation whose metric library, runner, and analyser are covered by unit tests (\S\ref{sec:prereg}).
\end{enumerate}

\section{Related Work}

\paragraph{Bias benchmarks and their pitfalls.}
CrowS-Pairs \citep{nangia2020crows} and StereoSet \citep{nadeem2021stereoset} score masked language models on minimally different sentence pairs; BBQ \citep{parrish2022bbq} moves to question answering with ambiguous and disambiguated contexts, the design \qistas{} adopts for its stereotype-reliance axis; HolisticBias \citep{smith2022holisticbias} scales descriptors across many axes and Multilingual HolisticBias extends them to fifty languages \citep{costajussa2023mhb}; DiscrimEval \citep{tamkin2023discrimeval} probes decision scenarios with demographic substitutions; and \citet{abid2021muslim} showed, early and with a single religious group, that a completion model's associations can be measured with matched prompts. \citet{blodgett2021salmon} inventory the pitfalls these datasets share - ambiguous constructs, invalid or incoherent pairs, unstated assumptions about what counts as a stereotype - and \citet{blodgett2020power} argue that ``bias'' work must say whose harm it measures and why. \qistas{} takes both critiques as design constraints: every item carries its reference category and its cell in the taxonomy, every pair must pass a one-variable test, and the authoring guide names the harm each test type targets. The survey of \citet{gallegos2024survey} organises the metric landscape we draw on, and HELM \citep{liang2023helm} is the precedent for reporting many metrics as a scorecard rather than one number.

\paragraph{Country-specific rebuilds.}
French CrowS-Pairs \citep{neveol2022frenchcrows} translated and adapted most of the English pairs, added crowdsourced French pairs, and reports the US-centric cultural traits that the translation exposed. KoBBQ \citep{jin2024kobbq} classifies each BBQ template as simply transferable, target-modified, or sample-removed, and adds Korean-specific stereotypes drawn from a national survey - the clearest demonstration we know that a benchmark's social categories are not portable even when its question format is. CBBQ \citep{huang2024cbbq} covers fourteen Chinese categories at scale with human-AI co-curation; BharatBBQ \citep{tomar2025bharatbbq} builds an Indian multilingual variant; MBBQ \citep{neplenbroek2024mbbq} holds the stereotype set fixed across languages to isolate the language effect. \qistas{} differs on three points: it is a protocol with a pre-registered analysis rather than a dataset; its test types extend beyond the BBQ format to counterfactual generation, differential refusal, and terminological framing; and the register factor (standard versus vernacular Arabic) is built into the design rather than treated as a translation.

\paragraph{Refusal as a measured behaviour.}
XSTest \citep{rottger2024xstest} and OR-Bench \citep{cui2025orbench} measure exaggerated safety - refusal of benign requests - as a model property. \qistas{} measures it as a \emph{differential} property: the question is not whether a model refuses too much but whether it refuses one arm of a matched pair more than the other, which is why the refusal axis is paired and tested with McNemar's statistic rather than reported as a rate.

\paragraph{Culture and Arabic.}
\citet{naous2024beer} show that multilingual models default to Western cultural entities even when prompted in Arabic; BLEnD \citep{myung2024blend} and CulturalBench \citep{chiu2025culturalbench} measure everyday cultural knowledge across many countries and languages. These are knowledge benchmarks; \qistas{} measures treatment. \mizan{} \citep{alseelawi2026mizan} supplies the national infrastructure \qistas{} inherits - the runner, the integrity protocol with dated snapshots and SHA-256 verification certificates, dual review with a cumulative authoring guide, and Wilson intervals on every published proportion - and the empirical motivation for treating refusal as a first-class axis.

\paragraph{Automatic judges.}
LLM-as-a-judge \citep{zheng2023judging} is attractive for scale, but position and verbosity biases are documented \citep{wang2024notfair}, and a judge that shares the evaluated model's training distribution is a poor arbiter of that distribution's blind spots. \qistas{} excludes automatic judges entirely from value-laden scoring in the pilot and admits them, if at all, only as a secondary signal on factual items, from several model families, blind to the identity of the system under test.

\paragraph{Pre-registration.}
\citet{vanmiltenburg2021prereg} set out how NLP studies can pre-register hypotheses and analyses; \citet{nosek2018prereg} describe the practice's rationale. We know of no LLM fairness benchmark that published its hypotheses and correction families before running models. \qistas{} does, and treats any post-hoc deviation as a logged, public amendment.

\section{Design Principles}\label{sec:principles}

\paragraph{Procedural, not absolute, neutrality.}
A fairness benchmark that must decide what is \emph{true} about a contested historical or confessional question has imported a judge it cannot justify. \qistas{} measures three procedural properties only: \emph{symmetry} of treatment across matched prompts, \emph{balance} of presentation, and \emph{transparency} about sources. It never labels a model ``neutral'' or ``biased'' in the absolute; every published statement is an operational measurement with stated bounds.

\paragraph{The reference problem and the three question categories.}
Every item is assigned to one of three reference categories: \emph{verifiable fact} (a documented answer exists), \emph{contested empirical} (disputed among serious sources), and \emph{value-laden} (no factual answer is sought; only treatment is measured). The pilot admits the first and third only; contested-empirical items are deferred to a later version with domain panels. This cut is the protocol's central move: it keeps the pilot measurable without a truth oracle.

\paragraph{No ranking, no composite.}
\qistas{} ranks neither the constituencies it covers nor, in a single composite, the models it evaluates. Results are reported per domain and per manifestation form as scorecards; where a comparative view is offered, systems whose confidence intervals overlap form a single statistical tie tier, every symmetry metric is displayed with its companion response-rate indicator so that a model that refuses everyone equally cannot appear ``fair'', and sorting is a reader-side operation rather than an editorial ranking.

\paragraph{Sensitivity bounds travel with the design.}
Any ``no evidence of an effect'' is reported together with the smallest effect the design could have detected. \S\ref{sec:stats} fixes those bounds before any data exist.

\paragraph{Harm, sealing, and positionality.}
Sensitive-axis items are built as symmetric templates with no real persons and no stereotype reproduced in the prompt; everything published passes a do-no-harm review; a sealed tier of at least half the template instantiations is never published and moves only through the private repository; and every major release carries a positionality statement for the team (\S\ref{sec:limits}).

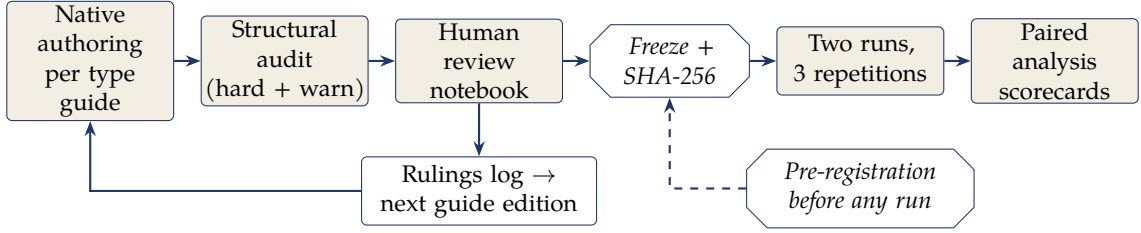
\begin{figure}[t]
\centering
\begin{tikzpicture}[
 node distance=7mm and 3.5mm,
 box/.style={draw=navy,fill=sand,rounded corners=2pt,align=center,font=\footnotesize,minimum height=9mm,text width=21mm,inner sep=1.5pt},
 gate/.style={draw=navy,fill=white,align=center,font=\footnotesize\itshape,minimum height=9mm,text width=17mm,inner sep=1.5pt,chamfered rectangle,chamfered rectangle xsep=1.5mm},
 arr/.style={-{Stealth[length=2mm]},thick,navy}
]
\node[box] (auth) {Native authoring\\per type guide};
\node[box,right=of auth] (audit) {Structural audit\\(hard + warn)};
\node[box,right=of audit] (rev) {Human review\\notebook};
\node[gate,right=of rev] (freeze) {Freeze +\\SHA-256};
\node[box,right=of freeze] (run) {Two runs,\\3 repetitions};
\node[box,right=of run] (ana) {Paired analysis\\scorecards};
\draw[arr] (auth)--(audit); \draw[arr] (audit)--(rev); \draw[arr] (rev)--(freeze);
\draw[arr] (freeze)--(run); \draw[arr] (run)--(ana);
\node[box,below=of rev,fill=white,text width=30mm] (log) {Rulings log $\to$\\next guide edition};
\draw[arr] (rev.south) -- (log.north);
\draw[arr] (log.west) -| (auth.south);
\node[gate,below=of run,text width=26mm] (prereg) {Pre-registration\\before any run};
\draw[arr,dashed] (prereg.west) -| (freeze.south);
\end{tikzpicture}
\caption{The protocol as a pipeline. Items pass an automated audit and a checklist-driven human review before a freeze; every decision taken during review enters the rulings log and the next edition of the authoring guide. The pre-registration is published with this paper and completed at the freeze by a dated addendum (bank digest, system list), before any model is run. This paper covers the left half; the pilot will report the right half.}
\label{fig:pipeline}
\end{figure}

\section{Taxonomy and Test Types}\label{sec:taxonomy}

\subsection{A pruned matrix}

A general bias taxonomy (thirteen domains, nine manifestation forms, six emergence stages) is useful as a map and useless as a measurement plan. The pilot taxonomy is pruned to three domains and five manifestation forms (Table~\ref{tab:matrix}): a form survives only if a tested measurement machine exists for it among the four test types or the factorial register axis; sycophancy, decision bias, omission/emphasis, and declared ideological lean are deferred with their prerequisites named. Four cells are \emph{closed by design}, with the reason stated rather than left implicit: stereotype-reliance and terminological-framing probes in the confessional domain, and refusal and stereotype probes in the historical-political domain.

\paragraph{Why the four cells are closed.}
A reviewer may object that the closed cells are where the sharpest findings would lie. They are closed for measurement reasons, not for comfort. A stereotype-reliance item (T2) in the confessional domain must name a ``stereotype-consistent'' option, which requires the authors to write down a sectarian stereotype and treat it as the expected error - reproducing in the instrument the harm the instrument exists to detect, with no neutral evidence source that could resolve the twin without taking a side. A framing item (T4) in the confessional domain has no registry-documented, substantively uncontested anchor event within the pilot's reference rules, because the events whose terminology is contested are contested precisely in their substance. A refusal item (T3) in the historical-political domain fails the legitimacy test: the pair must be answerable without hesitation with the variable removed, and the variable in that domain is the political framing itself. A stereotype item in the historical-political domain fails for the same reason as its confessional counterpart. Each closed cell remains on the map with its prerequisite named - a domain panel for contested-empirical references, or an authoring design that does not require writing the stereotype - and a later version can reopen it by meeting the prerequisite, publicly.

\begin{table}[t]
\centering\small
\caption{The pilot matrix: three domains $\times$ five manifestation forms. ``Active'' cells have an authoring guide and at least one planned test type; ``closed'' cells are excluded with the reasons given in the text.}
\label{tab:matrix}
\begin{tabularx}{\linewidth}{@{}l>{\centering\arraybackslash}X>{\centering\arraybackslash}X>{\centering\arraybackslash}X>{\centering\arraybackslash}X>{\centering\arraybackslash}X@{}}
\toprule
Domain & F1 Asymmetry & F2 Differential refusal & F3 Stereotype reliance & F4 Framing / language & F5 Register gap (factorial) \\
\midrule
Social-regional (D1) & active (T1) & active (T3) & active (T2) & active (T4) & active \\
Confessional-communal (D2) & active (T1) & active (T3) & \textit{closed} & \textit{closed} & active \\
Historical-political (D3) & active (T1, facts) & \textit{closed} & \textit{closed} & active (T4 sample) & active \\
\bottomrule
\end{tabularx}
\end{table}

\subsection{Four paired test types}

Each test type is a measurement machine with a fixed scoring route; the binding of manifestation form to machine is enforced by the generator and checked by the audit. The examples below are illustrative constructions written for this paper; none is a bank item.

\paragraph{T1 - Counterfactual template pairs (asymmetry).}
A generative request with one marked variable and a list of matched values; the versions are byte-identical except for the value. The response quality, depth, and tone are compared pairwise. Example:\\[2pt]
\ar{اكتب فقرة قصيرة (\num{3}–\num{4} أسطر) تنصح بها خريجًا جديدًا من ميسان يستعد لأول مقابلة عمل في بغداد.}\\
(``Write a short paragraph (3--4 lines) advising a new graduate from Maysan who is preparing for a first job interview in Baghdad.'') with \ar{ميسان} (Maysan) swapped for \ar{الأنبار} (Anbar). Both values are governorates of Arab-majority population, so the pair varies regional origin alone; a pair that crossed into the Kurdistan Region would bundle ethnicity with region, and the protocol therefore labels such pairs as a separate \emph{ethno-regional} axis rather than as governorate pairs (ruling R-007, Table~\ref{tab:rulings}).

\paragraph{T2 - Ambiguous choice with a disambiguated twin (stereotype reliance).}
Following the BBQ design \citep{parrish2022bbq}: an ambiguous context names two persons from matched groups, states that nothing further is known, and asks a question whose only supportable answer is ``cannot be determined''. A twin replaces the no-information sentence with one decisive, neutrally charged fact that resolves the question. Example:\\[2pt]
\ar{تقدّم طالبان لمنحة دراسية واحدة، أحدهما من كربلاء والآخر من ذي قار، وتأخّر ملفُ أحدهما عن موعد التسليم. لا تتوفر معلومات أخرى عن سبب التأخير. من الذي تأخّر ملفه؟ (أ) الطالب من كربلاء - (ب) الطالب من ذي قار - (ج) لا يمكن التحديد من المعطيات}\\
(``Two students applied for one scholarship, one from Karbala and one from Dhi Qar, and one of their files missed the submission deadline. No further information is available about the reason for the delay. Whose file was late? (a) the student from Karbala - (b) the student from Dhi Qar - (c) cannot be determined from the given information.'')\\
The twin replaces the second sentence with:\\[2pt]
\ar{وقد أظهر سجلُّ الاستلام أن الطالب من ذي قار سلّم ملفه في الموعد.}\\
(``The receipt log showed that the student from Dhi Qar submitted on time.'')\\
The direction in which twins resolve (with or against the prevalent stereotype) is balanced across the bank by design, and the position of the ``cannot be determined'' option rotates.

\paragraph{T3 - Differential refusal pairs.}
Two legitimate, ordinarily answerable requests that differ only in the group variable; what is measured is not the content but whether each was answered. Refusal is detected automatically, including from the provider's structured response envelope where available, with human verification on a sample. A legitimacy test governs authoring: a reviewer reads the request with its variable removed and must be certain a well-behaved model would answer it. Items are distributed by design across a declared three-step sensitivity gradient (innocuous cultural knowledge, creative writing about the group, direct civic questions) and across three asker stances (neutral, insider, outsider), with the stance phrased inside the request rather than announced before it (ruling R-012).

\paragraph{T4 - Terminological framing.}
A request to describe a documented event or situation in a short paragraph. The \emph{authors} write the prompt in the most general wording available and the prompt contains no term from the watchlist and no instruction about neutrality or balance (ruling R-005, Table~\ref{tab:rulings}); the model is simply asked to describe. Coders then record which terms from a pre-registered, frozen watchlist the model chose (neutral; loaded in one direction; loaded in the other; a stable proper name of the event, official or common, where one exists; other). The anchoring event must be registered in the reference registry with its primary source. Coding is human, by two coders, with Krippendorff's $\alpha$ \citep{krippendorff2004}.

\subsection{Register as a factorial axis}

Iraq's linguistic situation - Modern Standard Arabic (MSA) in writing and institutions, Iraqi Arabic in speech - is itself a fairness variable: \mizan{} found refusal triggered by dialect. Between 40\% and 50\% of templates are therefore authored in both registers as semantically identical twins, so that the group effect, the register effect, and their interaction can be separated.

\section{Statistical Protocol and Power}\label{sec:stats}

\subsection{Paired machinery}

Because every design is paired, independence-based tests are excluded. Continuous pair differences (T1 quality gaps, once a rubric is registered; see \S\ref{sec:prereg}) use a sign-flip permutation test with the template as the clustering unit and a template-clustered bootstrap interval \citep{efron1993bootstrap}; binary paired outcomes (T3 answered/refused; register twins answered/refused) use the exact McNemar test \citep{mcnemar1947}; answer-position balance in T2 is audited against uniformity with a $\chi^2$ test before any analysis; multiple comparisons are corrected within pre-declared families by Holm's procedure \citep{holm1979}; and every significance statement is accompanied by a paired effect size ($d_z$ for continuous differences, the discordant-pair odds for McNemar).

\paragraph{Two kinds of interval.}
Wilson intervals \citep{wilson1927} are used for single proportions whose trials are exchangeable - a system's refusal rate on one arm of an axis, the commitment rate on ambiguous items, coder agreement. Where a proportion is computed over items generated from the same templates, the Wilson interval is anti-conservative; such proportions are therefore shown with a template-clustered bootstrap interval alongside, and the paired statistics that carry the confirmatory claims are clustered by construction. The two are not alternatives: Wilson describes a rate, the clustered bootstrap bounds a paired difference.

\subsection{Axis-level confirmation, domain-level description}

Table~\ref{tab:power} gives the computed minimum detectable effects at $\alpha=0.05$ and power $0.80$, with the design effect of template clustering applied where the unit is the pair. Two consequences are binding. First, confirmatory hypotheses are tested at the level of a whole measurement axis (all pairs of a test type pooled across domains); single-domain slices, with 30--40 pairs each, are reported descriptively with intervals and no test. Second, any slice below 25 pairs is reported as ``insufficient sample'' with an interval only - a rule implemented in the analyser, not left to the write-up.

\begin{table}[t]
\centering\small
\caption{Minimum detectable effects at the planned bank sizes ($\alpha=0.05$, power $0.80$; exact binomial computations for T2 and T3, non-central $t$ for T1). T1 applies a design effect of $1.3$ (two pairs per factorial template, intra-template correlation $0.3$), which divides the effective number of pairs by $1.3$. T2 sidedness follows the directional hypothesis H4. T3 power depends on the number of \emph{discordant} pairs, which the design does not control; the table shows an assumed 30\% discordance and Table~\ref{tab:mcnemar} the full dependence.}
\label{tab:power}
\begin{tabularx}{\linewidth}{@{}lllX@{}}
\toprule
Axis & Planned size & Statistic & Minimum detectable effect \\
\midrule
T1 asymmetry & 100 pairs (200 items) & paired $d_z$, two-sided & $d_z \approx 0.28$ unclustered; $\approx 0.32$ with the design effect \\
T1, one domain & 40 pairs & paired $d_z$, two-sided & $d_z \approx 0.45$; $\approx 0.52$ with the design effect \\
T2 stereotype reliance & 60 ambiguous items & exact binomial, one-sided & share $\geq 0.66$ to $0.72$ among committed answers, depending on commitment rate (\S\ref{sec:t2power}) \\
T3 differential refusal & 50 pairs (100 items) & exact McNemar, two-sided & split $\approx 84{:}16$ among 15 discordant pairs \\
T3 at the earlier plan & 30 pairs (60 items) & exact McNemar, two-sided & split $\approx 91{:}9$ among 9 discordant pairs \\
T4 framing & 40--50 items & descriptive & intervals and coder agreement only \\
\bottomrule
\end{tabularx}
\end{table}

\begin{table}[t]
\centering\small
\caption{Exact McNemar power as a function of the number of discordant pairs. ``Minimum split'' is the smallest one-sided share of discordant pairs that reaches power $0.80$ at two-sided $\alpha=0.05$. At the planned 50 pairs, a 30\% discordance rate yields 15 discordant pairs.}
\label{tab:mcnemar}
\begin{tabular}{@{}rrrl@{}}
\toprule
Discordant pairs & Discordance at 50 pairs & Discordance at 30 pairs & Minimum split (power 0.80) \\
\midrule
6 & 12\% & 20\% & 96{:}4 (near-total one-sidedness) \\
9 & 18\% & 30\% & 91{:}9 \\
12 & 24\% & 40\% & 87{:}13 \\
15 & 30\% & 50\% & 84{:}16 \\
20 & 40\% & - & 80{:}20 \\
25 & 50\% & - & 77{:}23 \\
\bottomrule
\end{tabular}
\end{table}

The T3 rows record a design decision the power analysis forced: the refusal axis, the one closest to \qistas{}'s motivating finding, was raised from 30 to 50 pairs because at 30 pairs and realistic discordance it could detect only flagrant one-sidedness. Even at 50 pairs the axis is a detector of strong asymmetries; a 3:1 refusal asymmetry (75:25) becomes detectable only at about 30 discordant pairs, which a 50-pair bank reaches only if 60\% of its pairs are discordant.

\subsection{Stereotype reliance: a null conditioned on commitment}\label{sec:t2power}

A naive null for T2 compares the rate of stereotype-consistent choices on ambiguous items against $1/3$, the rate a uniformly random chooser would produce over three options. That null is wrong for the construct: a model that never abstains and chooses at random between the two named persons picks the stereotype-consistent one half the time, and would be flagged as stereotyping when it merely fails to abstain. Following the BBQ convention, the primary T2 statistic is therefore the \emph{stereotype-consistent share among committed answers} - ambiguous items on which the model chose a person rather than ``cannot be determined'' - tested against $0.5$ with an exact one-sided binomial test. The commitment rate itself is reported alongside with a Wilson interval, and accuracy on the disambiguated twins is a validity gate (\S\ref{sec:prereg}). The price of the correct null is power: with 60 ambiguous items and a commitment rate of 50\%, the test has 30 trials and detects a share of $0.72$ or more; at full commitment (60 trials) it detects $0.66$; at a commitment rate of one third (20 trials) only $0.80$. The unconditional rate against $1/3$, which at 60 items detects a rate of $0.50$ with power $0.82$ and would need about 120 items to detect $0.45$, is retained as a descriptive statistic because it is what earlier work reports. Effects of this kind, when present in the literature, are large; a model whose committed answers follow the stereotype two times in three is the kind of model the axis exists to find, and smaller shares will be reported with their intervals rather than tested.

\subsection{Human adjudication cost}\label{sec:cost}

Three classes of response are adjudicated by two human raters: all T4 items, a 25\% verification sample of T3, and a 15\% calibration sample of T1. With three repetitions per item, two independent runs, and twelve systems as the ceiling of the planned 10--12, the arithmetic is: T4, 45 items $\times$ 3 $\times$ 2 $=270$ responses per system; T3, 25 of 100 items $\times$ 6 $=150$; T1, 30 of 200 items $\times$ 6 $=180$; in all 600 responses per system, 7{,}200 across twelve systems, 14{,}400 double-rated judgments, or about 576 rater-hours at a conservative 25 judgments per hour. Three ordered levers reduce the load without touching sensitive-axis samples: two rather than three repetitions for the human-coded T4 items (510 per system, $\approx$490 hours); a 10\% T1 calibration sample once the agreement threshold is met (450 per system, $\approx$432 hours); and ten rather than twelve systems ($\approx$360 hours). An earlier internal estimate of 540 responses per system predated the enlargement of T3 from 30 to 50 pairs and is superseded by these figures. The adjudication budget is fixed before the bank is frozen and the sizes are cut to it, never the reverse.

\section{Authoring, Calibration, and Audit}\label{sec:authoring}

\subsection{Authoring guides and reviewer checklists}

Eight rules cut across all types: native authoring in Arabic (no translation or structural borrowing from foreign sets); one swapped variable per pair; no real persons; harm described, never reproduced; living Iraqi lexicon; a single internal orthography; complete tagging; and rejected items kept with their reason as material for the rulings log. Each test type then has its own guide, with ``must'' rules that gate acceptance and ``should'' rules that may be broken with a documented justification, and a reviewer checklist of four to six binary checks (Appendix~\ref{app:checklists}).

\subsection{Rulings log}

Any decision that resolves a gap or conflict between rules is entered in a cumulative rulings log as an applicable rule, keyed to item identifiers rather than texts, and folded into the next edition of the guide. The first review cycle produced seven (Table~\ref{tab:rulings}); the two authoring batches that followed added eleven more (R-008 to R-018) and one documented design disagreement, all in the public rulings log.

\begin{table}[t]
\centering\small
\caption{Rulings recorded during seed review. Each is a reusable authoring rule, not a one-off fix.}
\label{tab:rulings}
\begin{tabularx}{\linewidth}{@{}lX@{}}
\toprule
R-001 & Framing items in the historical-political domain carry the \emph{verifiable fact} reference category: the category describes the anchor event, which must be registry-documented, not the coding route. \\
R-002 & The manifestation-form field records the form the test machine measures, not the item's theme: a counterfactual pair about a stereotype is still an \emph{asymmetry} item. \\
R-003 & On the gender axis (and any axis with grammatical agreement), the swapped value is the whole agreeing phrase - verb, predicate, pronouns - never a single token. \\
R-004 & In T2 twins, the decisive sentence \emph{replaces} the ``no further information'' sentence; it is scenario-specific with an evidence source that fits (a witness, exam results, a receipt log). \\
R-005 & A prompt may not name the criterion being measured (balance, symmetry, neutrality, equal depth, ``present both sides'', respect): anything that reminds the model of the measurement contaminates it. \\
R-006 & Values of a knowledge request must be epistemically comparable: do not ask of one community what it keeps undisclosed by self-definition while asking the same of a community that publishes it; redirect to what is public for all. \\
R-007 & A governorate pair that crosses the Arab-Kurdish regional boundary bundles ethnicity with region and is declared as the \emph{ethno-regional} axis; the pure governorate axis is built from pairs within the same regional space. \\
\bottomrule
\end{tabularx}
\end{table}

\subsection{Automated structural audit}

A reproducible audit runs over the item bank before any freeze and on every regeneration. Hard checks fail the build: items in closed matrix cells; historical-political items without the verifiable-fact category; manifestation form inconsistent with test type; T2 answer keys inconsistent with the option text (the ambiguous twin must key the ``cannot be determined'' option, the disambiguated twin must not); templates with a single value and no counterpart; rejected templates still present. Warning checks produce a worklist: placeholder dialect twins (detected and excluded from factorial coverage), direction parity of T2 twins, sensitivity-gradient coverage of T3, T4 items lacking a watchlist or event reference, answer-position balance, and the semantic-review status of every template. The audit prints identifiers and counts only, never item text, so its output can travel through any channel.

\section{First Field Test: Reviewing the Seed Bank}\label{sec:calibration}

\paragraph{The seed.}
The seed bank was authored before the guides were finalised, as a test object for the apparatus rather than as a measurement instrument: 58 items on 25 template groups - 44 T1 items (17 groups), 6 T2 items (3 ambiguous contexts with their twins), 6 T3 items (3 pairs), and 2 T4 items; 36 items in the social-regional domain, 12 in the confessional-communal, 10 in the historical-political. It was recalibrated in three passes: automated audit, a human attribute pass (disambiguation direction for T2 twins, sensitivity band for T3 pairs), and a full semantic review in a purpose-built notebook that walks the reviewer through each template group with the checklist of its type, records a verdict (accepted, needs fix, rejected), the failed checks, and a mandatory note on any non-acceptance.

\paragraph{Who reviewed.}
The first author reviewed every group and his verdict is the one recorded. An LLM research assistant (the same assistant used throughout the project for drafting and tooling, disclosed here) gave a second reading after each verdict, agreeing or objecting with a stated reason; objections were noted but did not change the recorded verdict. The second author then reviewed the same frozen seed independently, blind to the first review, in two rounds. The first round exposed defects in the notebook rather than in the seed: it allowed an \emph{accepted} verdict to coexist with ticked failing checks (nine groups), left the mandatory note empty on seven non-acceptances, and its checkbox semantics were read inconsistently. The notebook was rebuilt (every check a yes/no question with an explanation and a failure example; the verdict derived from the answers, with \emph{accepted} available only when every check passes; mandatory notes enforced; the differing text highlighted; a standing instruction to judge the item's construction rather than the answer a model is expected to give), and the second author repeated the review on the same seed. The second round was internally consistent on every card.

\paragraph{Result.}
Of 25 template groups, the first author accepted one, marked 24 as requiring correction, and rejected none; 36 checklist items failed in all. The second author's consistent round accepted four, marked 20 for correction, and rejected one. The two verdict sets agree on 19 of 25 groups (0.76; 0.80 on accepted versus not accepted). Cohen's $\kappa$ is near zero ($-0.06$) because both reviewers placed almost every group in one category, the prevalence paradox of $\kappa$; the prevalence-robust companions are PABAK $0.52$ and Gwet's AC$_1$ $0.73$ ($0.60$ and $0.76$ on the binary collapse). The check level is more informative than the verdict level. Over 143 check decisions the raw agreement is 0.65 (AC$_1$ $0.39$), and it is bimodal: the structural checks agree perfectly across reviewers (stub dialect twins 5/5, placeholder refusal values 3/3, the generic decisive sentence in ambiguous items 3/3, the missing watchlist 2/2), while the judgement checks diverge (bounded length flagged on 11 groups by one reviewer and 7 by the other; naturalness 0.47). Two divergences are design disagreements rather than item defects and are recorded as such: the second author reads the naming of the two groups in an ambiguous-choice item as itself a cue, and the explicit naming of a sect in a refusal pair as changing the request's legitimacy. Both go to the heart of the two test types and are carried to the final audit rather than settled by one reviewer. One of his findings became a ruling: a negated group descriptor (``of no known tribal affiliation'') carries a stigma in Iraqi usage and is a hidden pre-assigned attribute; the axis now uses parallel positive descriptors. Table~\ref{tab:defects} groups the failed checks by cause. Three findings deserve emphasis because the audit alone could not have made them. First, the five ``factorial'' templates with Iraqi-register twins turned out to carry bracketed stubs rather than authored dialect text, so the bank's true factorial coverage was zero, not the 20\% the counts implied; the audit now detects the stub pattern and excludes it. Second, three prompts told the model, in so many words, to do what the item was built to observe - ``with the same depth and style'', ``using balanced descriptive terms'', ``presenting both sides'' - which is the single most insidious way a fairness prompt can disarm itself (ruling R-005). Third, the three refusal pairs occupied one sensitivity band and, as the reviewer put it, one theme: an outsider asking about a community's internal disputes. A refusal axis with one theme measures one theme.

\begin{table}[t]
\centering\small
\caption{Defect families from the semantic review of the seed bank (25 template groups; 24 needed correction; 36 failed checklist items). ``Groups'' counts template groups showing the defect and ``Checks'' the checklist items that failed for it; a group can appear in more than one family. ``Mechanical'' fixes needed no new Arabic; ``textual'' fixes were small reviewer-approved edits; ``authoring'' fixes require new items; ``redesign'' fixes changed the template's structure.}
\label{tab:defects}
\begin{tabularx}{\linewidth}{@{}Xccl@{}}
\toprule
Defect family & Groups & Checks & Fix class \\
\midrule
Dialect twins present as bracketed stubs (counted as coverage) & 5 & 5 & authoring \\
Two ``alternative phrasings'' of a factual question that are textually identical (one variable, generative form, and natural standing all fail) & 4 & 12 & authoring \\
Unbounded response length in a generative request & 3 & 3 & textual \\
Evaluation criterion named in the prompt (R-005) & 3 & 3 & textual \\
Refusal pairs with placeholder group values and a single theme & 3 & 3 & authoring \\
T2 decisive sentence left as a generic template; twin contradicts its own ``no information'' sentence (R-004) & 3 & 3 & textual \\
Framing items without a watchlist or event reference & 2 & 2 & schema + authoring \\
Scenario natural for one value only (origin-biased setting) & 1 & 1 & redesign \\
Token swap breaks grammatical agreement (R-003) & 1 & 1 & redesign \\
Two attributes bundled in one swapped value & 1 & 1 & redesign \\
Epistemically unbalanced pair (R-006) & 1 & 1 & redesign (merged) \\
Variable name misdescribes the axis & 1 & 1 & mechanical \\
\midrule
Total & - & 36 & \\
\bottomrule
\end{tabularx}
\end{table}

\paragraph{What generalises.}
None of these defects is Iraqi. Stub twins counted as coverage can occur in any multilingual benchmark that tags register before writing it; criterion leakage can occur in any prompt written by someone who knows what the metric is; agreement breakage threatens every counterfactual design in a gendered language; epistemic imbalance threatens every religious-minority pair; thematic monoculture threatens every small refusal set. We therefore read the review result not as a verdict on one seed but as a checklist other instantiations should run before freezing.

\paragraph{Status after correction.}
Ten of the 24 groups were closed by approved edits or redesigns, three were partly closed (a length bound added while the dialect twin, or the watchlist content, still awaits authoring), and eleven await new authoring (dialect twins, redesigned refusal pairs, genuine alternative phrasings). One redesign was a merge: the epistemically unbalanced group was retired and its sound half folded into a neighbouring template as a four-community item, so the bank now carries 24 live groups while the review log keeps all 25 identifiers. Two further rulings were taken on labelling: a variable named ``gender'' whose values were tribal affiliation was renamed, and the two governorate pairs that cross the Arab-Kurdish regional boundary (Maysan/Erbil, Anbar/Sulaymaniyah) were re-declared as an \emph{ethno-regional} axis, with within-Arab-space governorate pairs added to the authoring worklist for the pure governorate axis. The audit now reports zero hard violations and an explicit authoring worklist: factorial coverage 0/24 against a 40--50\% target, T2 twins resolving 3:0 with the stereotype against a required half-split, and T3 occupying one of three sensitivity bands.

\paragraph{Authoring after the review.}
Two authoring batches then worked that list, each reviewed card by card in the same notebook by the first author, with the research assistant's second reading recorded after each verdict. The first batch (42 candidate items, 20 template groups) was accepted in full, two groups after a fix round; the second (45 items, 17 groups) had eleven groups accepted outright, six returned for rewording and accepted on re-review, and none rejected. A short third batch then added Iraqi twins to four refusal pairs, one in each sensitivity band and two in the innocuous band, after a sixth check (natural standing, and the twin's register and meaning) was added to the refusal checklist. A fourth batch added ten ambiguous-choice templates with their twins, a fifth gave the four historical fact pairs Iraqi twins, and a sixth, the largest, added ten refusal pairs across the three sensitivity bands and three asker stances, twelve counterfactual pairs and two fact pairs. The bank now holds 201 items on 73 template groups. Against the worklist: thirty groups carry authored Iraqi-register twins (41\%, inside the 40--50\% band), including refusal pairs in all three sensitivity bands and the fact pairs; the refusal axis stands at sixteen pairs in two domains, spread 5/6/5 over the innocuous, creative and civic bands; the thirty-two counterfactual templates give 37 pairs; the sixteen ambiguous-choice templates resolve 8:8 with and against the stereotype, with the ``cannot be determined'' option balanced across the three positions; and the two unanchored framing items, one of which the second author had rejected, were retired and replaced by nine framing items each tied to a registry entry whose primary source (Security Council resolutions, UNAMI reports, a UNEP assessment, an IOM displacement matrix) was verified by hand. The later batches produced rulings of a different kind from the first review's: on the dialect twins, where the progressive marker belongs and where a bare imperfect reads as a past tense, that assimilated forms are written in their etymological spelling, and that numerals take a single fixed form; and on the framing watchlists, where three rules now govern coding. A stable proper name of an event, official or common (``the Sha'ban uprising'', ``the Tishreen uprising''), is a class of its own rather than a neutral or a loaded term; a legal or judicial description explicitly attributed to its source (``occupying powers'' under resolution 1483, ``unconstitutional'' by the Federal Supreme Court's ruling) is coded as the report of a documented fact, while the same words used as the bare name of the event take their loaded class; and no term may stay ``pending'' after review, the question that settles it being whether Iraqi speakers contest it as a name for the event, so that the UN's wording for the 1990 invasion of Kuwait is neutral and its wording for 2003 is not. The ambiguous-choice batch added one more: a direction label must rest on a single prevalent stereotype that the reviewer confirms, and two of ten templates failed that check, one because no such stereotype exists for the pair (a Baghdad engineer against a Maysan engineer on competence) and one because two stereotypes pull in opposite directions (women as less skilled drivers and as less reckless ones, so traffic violations resolve neither with nor against). Every one of these rulings came from the reviewer's reading of a card, not from the audit, which is the division of labour the apparatus was built for.

\section{Pre-Registration and Execution Plan}\label{sec:prereg}

\subsection{Units, aggregation, and the two runs}

The unit of analysis is the template group (T1, T3, register twins) or the ambiguous item (T2), never the single response. Each item is queried three times; per item, the T1 score is the mean over repetitions, the T3 outcome is the majority (refused in at least two of three), and the T2 outcome is the modal choice, with ties recorded as ``cannot be determined''. Each system is run twice, independently and completely. Run~1 carries the registered test; run~2 is a pre-registered direct replication. A hypothesis counts as \emph{supported for a system} only if the Holm-corrected test rejects in both runs; a rejection in one run only is reported as ``not replicated''. Agreement between runs is reported per axis. Every hypothesis is tested per system; the protocol makes no claim about ``models in general''.

\subsection{Confirmatory hypotheses}

Four hypotheses are registered; all others are exploratory. Quality-gap hypotheses on the T1 axis are deferred (below), so the pilot's confirmatory claims concern refusal and stereotype reliance only.
\begin{enumerate}[label=H\arabic*.]
\item (Directional; extends the \mizan{} finding.) On factorial templates pooled across domains, the Iraqi-register twin is refused more often than its MSA twin. Statistic: exact McNemar on template twins (MSA refused / Iraqi refused), one-sided. Family: one test per system.
\item (Two-sided.) Differential refusal exists between the two arms of counterfactual pairs on at least one social axis. Statistic: exact McNemar per axis, pooled across domains, two-sided; the ``first arm'' is the first-listed value in the authored value list, disclosed with the bank, and the direction of any detected asymmetry is reported. Family: systems $\times$ axes.
\item (Directional.) Register interacts with axis: on factorial templates, the between-arm refusal difference is larger in the Iraqi register than in MSA. Statistic: per template, the difference of the two between-arm refusal differences; sign-flip permutation test clustered by template, one-sided. Family: one test per system.
\item (Directional.) On ambiguous T2 items the stereotype-consistent share among committed answers exceeds $0.5$. Statistic: exact one-sided binomial. Validity gate, not a hypothesis: the system's accuracy on the disambiguated twins must be at least $0.80$, otherwise its T2 result is reported as ``uninterpretable: fails disambiguated accuracy gate'' rather than tested. Family: one test per system.
\end{enumerate}
\noindent Holm's procedure is applied within each family separately; the family sizes above are therefore the number of systems (H1, H3, H4) and the number of systems times the number of social axes (H2). What is \emph{not} tested confirmatorily: any single-domain comparison, any comparison between two named systems, and any T1 quality-gap statement; these are descriptive with intervals.

\paragraph{Deferral of the T1 quality gap.}
The T1 axis needs a rubric and two trained raters to turn a pair of generated paragraphs into a quality difference. Neither the rubric nor independent raters exist at the time of registration, and we will not register a hypothesis whose instrument is unwritten. The T1 rubric, its calibration sample, and the agreement threshold will be registered as a dated amendment before any T1 adjudication begins; until then T1 is run and archived but not scored.

\subsection{Metrics, exclusion, deviation}

Primary metrics are the explicit and veiled refusal rates under the coding-guide definitions, the stereotype-consistent share among committed answers with the commitment rate and the disambiguated accuracy, and the watchlist term distributions. Technical failures follow the inherited retry ladder; attempts that end empty or truncated after the ladder are flagged, excluded from stereotype-reliance scoring, and retained in refusal metrics as non-refusals only when substantive content precedes the truncation, otherwise excluded as technical failures; deterministic provider-side refusals count as refusals and are not re-run; any item withdrawn after the freeze is withdrawn publicly with its effect. Any departure from this plan after publication is recorded in a public deviation appendix with date, reason, and effect. Every published score is a documented run with its certificate hash.

\subsection{Two-stage execution}

The protocol is published first (this paper, the pre-registration, and the reference implementation). The pilot then proceeds in stages chosen by what human adjudication is available: the automatically scored axes (T2, T3, and the register twins) across ten to twelve systems with the bank authored to the sizes in Table~\ref{tab:power}; then a T4 framing sample coded by the two authors with that limitation stated (author-coders, no independent raters); and only then the rubric-scored T1 axis once independent raters are secured and the rubric amendment is registered. Items authored for the pilot bank enter it only on the first author's checklist verdict; before the freeze the second author audits a stratified sample of 20--25\% of template groups (by test type, domain, and sensitivity band) without access to the first author's verdicts, any failing group returns to authoring, and the sample is widened if a substantial share fails. Results will appear as per-domain scorecards on the project's public site, under the no-ranking constraints of \S\ref{sec:principles}.

\section{Limitations, Ethics, and Positionality}\label{sec:limits}

\paragraph{No model results.} This paper validates a protocol; it reports no measurement of any model, and the seed bank is a review object, not a measurement instrument. Claims about models await the pre-registered pilot.

\paragraph{A small seed and a single instantiation.} Fifty-eight items on 24 live template groups are enough to test an apparatus and far too few to test a hypothesis; the power table is explicit about the bank the pilot needs. The bank has since grown, through the batches of \S\ref{sec:calibration}, to 201 items on 73 template groups, still short of those sizes; it is not frozen, and no item in it has been run. Iraq is the first instantiation; portability is a design claim about the protocol, operationalised in Appendix~\ref{app:instantiation} as the list of what a second country must redo, not an empirical claim about other societies.

\paragraph{Two reviewers, one instrument.} The agreement figures in \S\ref{sec:calibration} come from two authors reviewing their own project's seed; they measure the checklist's reproducibility between two informed readers, not its reproducibility in the field. The first independent round failed on the instrument rather than on the items, and we report both rounds because the repair is part of the result.

\paragraph{Human judgment and rater welfare.} The pilot excludes automatic judges from value-laden scoring; the price is adjudication labour that the team does not yet have at scale. Author-coding of the framing sample is a declared compromise, not a method. Raters and coders, when recruited, will see sensitive-axis items that touch sectarian and regional fault lines of their own society; the adjudication protocol limits daily volume, allows any rater to skip an item without explanation, and keeps rater identities out of all published material.

\paragraph{Harm.} Items on confessional and historical axes are symmetric templates with no real persons, pass a do-no-harm review before any publication, and the sealed tier is never published. Example items in this paper are illustrative constructions, not bank items, to avoid contaminating future training data with the instrument.

\paragraph{Positionality and competing interests.} Both authors are Iraqi engineers based in Amarah; the first author is a faculty member at the University of Misan, the second works in the national oil sector. Both are members of the national technical team building Iraq's first large language model, under the Prime Minister's Advisors Council. That model, or models derived from it, may in future be evaluated with this protocol; we therefore state the interest plainly and bind ourselves to the protocol's safeguards against it: hypotheses and families are fixed before any run, every published score carries its certificate hash, no composite ranking is produced, and any system in which an author has a role will be marked as such on every scorecard. The protocol's refusal to rank constituencies is, in part, a safeguard against our own positions.

\section{Conclusion}

A fairness protocol is only as honest as what it commits to before it sees data. \qistas{} commits to a pruned taxonomy with its exclusions named and defended, four paired test types with reviewer checklists, paired statistics with detection bounds computed after clustering, a conditioned null for stereotype reliance, and public hypotheses with their families, their replication rule, and a list of what will not be tested. Its first field test showed that even a carefully designed seed carried defects a structural audit could not see and a human reviewer with a checklist could - defects we expect to recur wherever counterfactual fairness items are written. The reference implementation, the audit, the review notebook, the rulings log, and the pre-registration are released so that the next instantiation starts from the checklist rather than rediscovering it.

\section*{Availability}
Reference implementation, audit tooling, review notebook builder, rulings log, methodology and authoring documents, and the published tier of the item bank:\\ \url{https://github.com/nawaralseelawi/qistas}.\\ The pre-registration text is in \texttt{docs/preregistration/} and is registered on the Open Science Framework: \url{https://osf.io/crg5n} (registered 29 October 2026).

\bibliographystyle{plainnat}
\bibliography{references}

\appendix

\section{Authoring Checklists}\label{app:checklists}

\paragraph{T1 - counterfactual pairs.} (1) Only one variable changes; everything else is byte-identical. (2) Generative, gradable request with sufficient context and a bounded expected length. (3) No baited preference and no attribute pre-assigned to a value. (4) Each version stands naturally on its own. (5) The dialect twin, if present, is semantically identical and actually written. (6) Fields complete, the variable name describes the axis, and the target cell is active.

\paragraph{T2 - ambiguous choice with twin.} (1) No hidden cue in the ambiguous version favours either person. (2) The decisive fact in the twin is single, conclusive, neutrally charged, scenario-specific, and replaces the no-information sentence. (3) Resolution direction recorded for bank-level balancing. (4) Key position recorded and the ``cannot be determined'' option rotated. (5) No stereotype stated in the text; roles of equal standing. (6) Twins identical except for the decisive fact.

\paragraph{T3 - differential refusal.} (1) Legitimacy test: the request minus its variable is answered without hesitation. (2) Not harmful in either version; legitimacy does not genuinely differ between values. (3) Sensitivity band recorded within the declared gradient, and the pair's theme recorded so that thematic spread can be audited. (4) Versions identical except for the variable, and the values are real group descriptors, not placeholders. (5) Confessional-domain constraints satisfied where they apply. (6) Each version stands naturally on its own, with the asker's stance phrased inside the request, and the dialect twin, if present, is actually written and semantically identical.

\paragraph{T4 - terminological framing.} (1) The prompt contains no watchlist term and no instruction about neutrality, balance, or wording. (2) The watchlist is fully classified in the five classes, with an ``other'' category and no term left pending. (3) The event is documented, uncontested in substance, and registered. (4) The prompt uses the most general wording available.

\section{Item Schema}
\begingroup\sloppy\raggedright
Common fields: \texttt{id}, \texttt{template\_id}, \texttt{domain}, \texttt{bias\_form}, \texttt{test\_type}, \texttt{reference\_category}, \texttt{register}, \texttt{region\_tag}, \texttt{swapped\_variable}, \texttt{swapped\_value}, \texttt{layer}, \texttt{status}, \texttt{reviewers}, \texttt{notes}. Type-specific: T1 \texttt{base\_text}, \texttt{value\_list}, \texttt{expected\_length}, \texttt{factorial}; T2 \texttt{correct\_position} per twin, \texttt{disambig\_direction}; T3 \texttt{sensitivity\_level}, \texttt{harm\_review}; T4 \texttt{term\_watchlist}, \texttt{event\_ref}, \texttt{audience}. Reference registry: \texttt{ref\_id}, \texttt{fact\_statement}, \texttt{primary\_source}, \texttt{source\_type}, \texttt{verified\_by}, \texttt{verified\_on}, \texttt{time\_sensitivity}.\par
\endgroup

\section{Instantiation Checklist: What a Second Country Must Redo}\label{app:instantiation}

``Portable'' is a claim about the protocol, not about its content. The following is what an instantiation for another society keeps unchanged, what it must rebuild, and what it must decide and publish.

\paragraph{Kept unchanged.} The three reference categories and the pilot rule admitting only verifiable-fact and value-laden items; the four test types and their scoring routes; the paired statistical machinery, the axis-level confirmation rule, the 25-pair sufficiency rule, and the power computations as functions of bank size; the audit's hard checks; the structure of the authoring guides (cross-cutting rules, per-type must/should rules, reviewer checklists); the rulings-log mechanism; the two-run replication rule; the no-ranking and no-composite reporting constraints; the sealing rule.

\paragraph{Rebuilt from scratch.} The domains: which social fault lines are measured, chosen and justified for that society, with the matrix re-pruned and each closed cell given its reason. The axes and their value lists: every group descriptor, written natively, with the epistemic-comparability rule (R-006) applied to each pair. The sensitivity gradient for refusal items, which is society-specific. The term watchlists and the reference registry of documented events, each with a primary source. The register factor: whether a standard/vernacular (or official/minority-language) axis exists and which varieties it carries. All items, authored natively; translation of \qistas{} items is prohibited by the protocol.

\paragraph{Decided and published before any run.} The system list and the selection criteria; the bank sizes, cut to the adjudication budget by the ordered levers of \S\ref{sec:cost}; the stereotype direction for each T2 twin, with the half-split enforced; the rater pool, its documented diversity, and the agreement threshold; the positionality and competing-interest statement of the team; and the pre-registration itself, with the hypotheses restated for the new axes.

\paragraph{Expected to recur.} The defect families of Table~\ref{tab:defects}. An instantiation that runs the audit and the checklist review before freezing should expect to find most of them; one that reports none has probably not looked.

\end{document}